\documentclass[pmlr,twocolumn,10pt]{jmlr} 

\usepackage{booktabs}
\usepackage{siunitx}
\usepackage{siunitx}
\usepackage{caption}
\usepackage{subcaption}
\usepackage{graphicx}
\usepackage{paralist} 
\usepackage{float}
\usepackage{enumitem}

\usepackage[switch]{lineno}

\theorembodyfont{\upshape}
\theoremheaderfont{\scshape}
\theorempostheader{:}
\theoremsep{\newline}

\jmlrvolume{LEAVE UNSET}
\jmlryear{2026}
\jmlrsubmitted{LEAVE UNSET}
\jmlrpublished{LEAVE UNSET}
\jmlrworkshop{Conference on Health, Inference, and Learning (CHIL) 2026} 

\title[Alzheimer’s Learning Platform for Adaptive Care Agents]{ALPACA: A Reinforcement Learning Environment for Medication Repurposing and Treatment Optimization in Alzheimer's Disease}

\author{%
\Name{Nolan Brady} \Email{nolan.brady@colorado.edu}\\
\addr University of Colorado Boulder, United States
\AND
\Name{Tom Yeh} \Email{tom.yeh@colorado.edu}\\
\addr University of Colorado Boulder, United States
}

\begin{document}

\maketitle

\begin{abstract}
Evaluating personalized, sequential treatment strategies for Alzheimer’s disease (AD) using clinical trials is often impractical due to long disease horizons and substantial inter-patient heterogeneity. To address these constraints, we present the \emph{Alzheimer’s Learning Platform for Adaptive Care Agents} (ALPACA), an open-source, Gym-compatible reinforcement learning (RL) environment for systematically exploring personalized treatment strategies using existing therapies. ALPACA is powered by the Continuous Action-conditioned State Transitions (CAST) model trained on longitudinal trajectories from the Alzheimer’s Disease Neuroimaging Initiative (ADNI), enabling medication-conditioned simulation of disease progression under alternative treatment decisions. We show that CAST autoregressively generates realistic medication-conditioned trajectories and that RL policies trained in ALPACA outperform no-treatment and behavior-cloned clinician baselines on memory-related outcomes. Interpretability analyses further indicated that the learned policies relied on clinically meaningful patient features when selecting actions. Overall, ALPACA provides a reusable in silico testbed for studying individualized sequential treatment decision-making for AD.

\end{abstract}

\begin{keywords}
Alzheimer's Disease, ADNI, Medical Simulation Environment, Disease Forecasting
\end{keywords}

\paragraph{Environment and Code Availability}
The ALPACA environment proposed in this paper is made available for use on GitHub \footnote{\nolinkurl{https://github.com/nolanrbrady/ALPACA-RL}} and as a pip package \footnote{\nolinkurl{https://pypi.org/project/ALPACA-DT-Sim/}}.

\paragraph*{Institutional Review Board (IRB)}
This study used de-identified data from the ADNI. \footnote{Data used in the preparation of this article were obtained from the Alzheimer’s Disease Neuroimaging Initiative database (adni.loni.usc.edu). ADNI investigators contributed to the ADNI design and data collection but did not participate in this analysis. A complete list of ADNI investigators is available at \nolinkurl{http://adni.loni.usc.edu/wp-content/uploads/how_to_apply/ADNI_Acknowledgement_List.pdf}.} Therefore, IRB approval was not required.

\section{Introduction}

The prevalence of AD increased by 147.9\% worldwide between 1990 and 2019 \cite{Li_Feng_Sun_Hou_Han_Liu_2022}. Despite this growth, identifying effective treatment regimens remains constrained by the slow, expensive, and ethically bound nature of clinical trials. These constraints are particularly limiting for personalized treatment strategies, in which clinicians must decide what to prescribe, when to initiate therapy, and how to adapt treatment as patient trajectories evolve over the years. Consequently, much of the AD research has focused on a narrow set of therapeutic targets, most notably amyloid-$\beta$ and tau, while many medication repurposing and combination strategies remain comparatively underexplored.

In silico experimentation offers a potential path forward by shortening iteration cycles and enabling large-scale hypothesis generation but hinges on the availability of physiologically plausible simulations that respond meaningfully to intervention decisions. Mechanistic models, such as ordinary differential equations (ODEs), provide interpretability but are often vulnerable to misspecification. In contrast, much of the deep learning literature emphasizes trajectory reconstruction or disease state classification rather than closed-loop simulations. Although offline reinforcement learning can optimize policies from retrospective data, it becomes unreliable for both policy improvement and evaluation when candidate actions fall outside the support of the original dataset, thus limiting its use cases \cite{Fujimoto2018OffPolicyDR}.    

These challenges are particularly pronounced in AD because of its underlying biology. Once believed to arise from a largely monolithic cause and therefore amenable to a single dominant treatment strategy, recent molecular and clinical stratification studies indicate the presence of biologically distinct AD subtypes with differential treatment sensitivities \cite{Tijms_Vromen_Mjaavatten_Holstege_Reus_van_der_Lee_Wesenhagen_Lorenzini_Vermunt_Venkatraghavan_et_al._2024}. Compounding this complexity, evidence suggests that treatment effects are also time-dependent, such that the timing of intervention can yield meaningfully different patient outcomes \cite{Amirrad_Bousoik_Shamloo_Al-Shiyab_Nguyen_Montazeri_Aliabadi_2017, Winblad_Wimo_Engedal_Soininen_Verhey_Waldemar_Wetterholm_Haglund_Zhang_Schindler_2006, Perneczky2025ClinicallyMB, Tarawneh2024TheSF, Sims2023DonanemabIE}. Together, these factors give rise to a combinatorial space of patient profiles, medication classes, and initiation windows that are infeasible to explore exhaustively through randomized clinical trials.

\begin{figure}[t]
    \centering
    \includegraphics[width=0.95\linewidth]{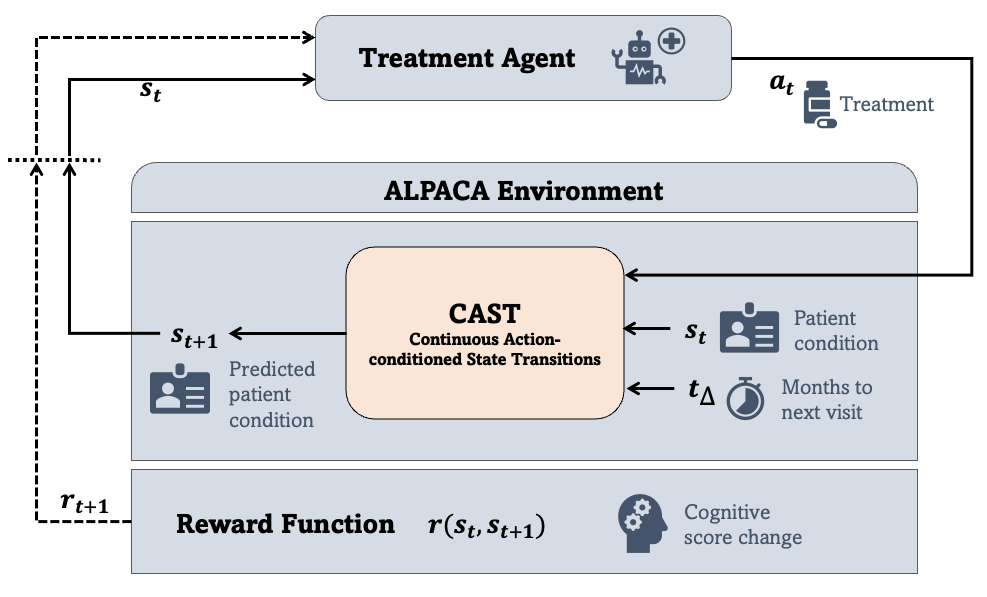}
    \caption{The ALPACA environment wraps the CAST model. Given a current state $s_t$ and medical action $a_t$, the CAST model predicts the future state $s_{t+1}$ ($t_{\Delta}$ months ahead), with $r_{t+1}$ representing the resulting reward. }
    \label{fig:architecture}
\end{figure}

To address these challenges, we introduce ALPACA (\emph{Alzheimer’s Learning Platform for Adaptive Care Agents}), an in silico reinforcement learning environment with continuous-valued clinical states, multi-binary medication actions, and transitions generated by CAST, a medication-conditioned autoregressive forecasting model trained on longitudinal trajectories from the ADNI dataset. Figure~\ref{fig:architecture} summarizes the ALPACA pipeline and CAST-based action-conditioned state transitions.

\noindent Our contributions are:
\begin{itemize}
    \item \textbf{CAST Model.} A medication-conditioned autoregressive model for multivariate AD trajectory simulation (Section 3.1).
    \item \textbf{ALPACA environment.} A Gym-compatible RL environment with continuous clinical states and a 17-class multi-binary medication action space (Section 3.2).
    \item \textbf{Benchmark analysis.} Policy learning and comparison with clinician baselines, accompanied by interpretability analysis (Section 4).
\end{itemize}

\section{Related Works}

To contextualize ALPACA, we reviewed two complementary areas: (i) simulation-based reinforcement learning in offline medical settings and (ii) prior AD-focused simulation environments.

\subsection{Simulation-based Reinforcement Learning Environments}

In many medical settings, reinforcement learning is primarily applied offline because of ethical and practical barriers to online experimentation \cite{Jayaraman2024APO}. Offline RL enables policy learning from static datasets \cite{Levine2020OfflineRL}, but it raises challenges for both policy optimization and evaluation when the target policy deviates from the data-generating distribution.

In offline reinforcement learning, the extrapolation error is the primary failure mode. Fujimoto et al.\ showed that the distributional mismatch between a learned policy and a fixed batch of data can lead to unreliable value estimates for underrepresented or unseen actions \cite{Fujimoto2018OffPolicyDR}. In clinical applications, constraining learned policies to support the observed clinician behavior can mitigate out-of-distribution recommendations and promote conservative improvements \cite{Huang2024SmartPR}. However, this constraint simultaneously limits a policy’s ability to generalize beyond its historical practice. The second major bottleneck lies in the evaluation of offline policies. When the evaluation policy deviates from the behavior policy, distribution shifts can induce high-variance estimates in off-policy evaluation (OPE), undermining reliable performance assessment \cite{Uehara2022ARO}. These challenges are further exacerbated in long-horizon settings, where compounding variance and bias make policy evaluation increasingly unstable \cite{Levine2020OfflineRL, Bossens2022LowVO}. Together, these limitations pose significant challenges in verifying the efficacy and safety of learned policies, which is critical in medical decision-making contexts.

Offline model-based reinforcement learning (offline MBRL) partially mitigates these issues by learning a transition dynamics model from retrospective data and using simulated rollouts for policy optimization. Model-based Offline Policy Optimization (MOPO), while not strictly medical, demonstrates the method’s ability to generalize past the bounds of static training data, where other offline RL policies cannot. MOPO achieves this by incorporating a penalty based on model uncertainty to manage distribution shifts during optimization while allowing controlled generalization beyond strict behavior cloning \cite{Yu2020MOPOMO}. The medical applications of offline MBRL include TR-GAN, which combines observational trajectories with simulated counterfactual rollouts for treatment recommendation \cite{Sun2022AdversarialRL}, and OMG-RL, which uses learned dynamics to support offline inverse reinforcement learning for dosing under a reward inferred from clinician behavior \cite{Lim2024OMGRLOfflineMG}. ALPACA extends the offline MBRL paradigm to AD by providing a medication-conditioned transition model in a reusable and Gym-compatible environment.

\subsection{Current State of AD Reinforcement Learning Environments}

Reinforcement learning has been applied across clinical domains such as sepsis, oncology, diabetes, and neurodegenerative disease to optimize sequential treatment policies \cite{Luo2024DTRBenchAI, Ghaffari2016AMR, Ahn2011DrugSO, Oberst2019CounterfactualOE, Man2014TheUT, Wang2023OptimizedGC, Bhattarai_Rajaganapathy_Das_Kim_Chen_Alzheimers_Disease_Neuroimaging_Initiative_Australian_Imaging_Biomarkers_and_Lifestyle_Flagship_Study_of_Ageing_Dai_Li_Jiang_et_al._2023}. 

In AD, only a limited number of studies have directly addressed the environment gap. Bhattarai et al.\ construct an environment based on a discretized Markov decision process derived from ADNI, in which Mini-Mental State Exam (MMSE) based cognitive measures are mapped to a finite set of disease stages using decision trees \cite{Bhattarai_Rajaganapathy_Das_Kim_Chen_Alzheimers_Disease_Neuroimaging_Initiative_Australian_Imaging_Biomarkers_and_Lifestyle_Flagship_Study_of_Ageing_Dai_Li_Jiang_et_al._2023}. The action space consists of a discrete set of interventions including AD medications, antihypertensives, and supplements \cite{Bhattarai_Rajaganapathy_Das_Kim_Chen_Alzheimers_Disease_Neuroimaging_Initiative_Australian_Imaging_Biomarkers_and_Lifestyle_Flagship_Study_of_Ageing_Dai_Li_Jiang_et_al._2023}. Although this formulation aligns naturally with value-based methods such as Deep Q-learning, it compresses continuous clinical trajectories into categorical disease states and substantially constrains the space of possible treatment strategies.

In contrast, Saboo et al.\ developed a mechanistic simulation based on coupled differential equations modeling amyloid pathology, brain atrophy, and neural activity \cite{Saboo_Choudhary_Cao_Worrell_Jones_Iyer_2021}. Reinforcement learning was used in this setting to learn the patterns of information processing load across simulated brain regions. Consequently, the learned policy functions primarily as a model of brain responses to pathology rather than as a framework for optimizing sequential treatment decisions \cite{Saboo_Choudhary_Cao_Worrell_Jones_Iyer_2021}.

These efforts advance AD simulation; however, common limitations persist. Existing environments often rely on coarse state abstractions and restricted medication spaces, limiting their ability to evaluate diverse treatment sequences and repurpose hypotheses. ALPACA addresses these gaps by modeling continuous clinical states and using autoregressive, medication-conditioned transitions that support counterfactual rollouts in a multi-binary action space spanning 17 therapeutic classes.

\section{ALPACA: Transition Model and Environment}
\label{sec:methods}
First, we introduce the CAST model and discuss its role as a forecasting engine for the ALPACA environment. Finally, we cover environmental assumptions and reward function design.

\subsection{CAST Model}

\subsubsection{Model Architecture}

The CAST model employs a mixture-of-experts (MoE) transformer architecture designed to capture the heterogeneity of the progression of Alzheimer’s disease \cite{DBLP:journals/corr/ShazeerMMDLHD17}. Unlike standard transformers, the MoE design allows distinct experts to specialize in different medical scenarios, resulting in higher-fidelity trajectory predictions. The network is composed of three transformer layers with an embedding dimension of 256 and four attention heads. In each layer, only one of the eight experts is activated per visit to ensure computational efficiency. The model operates in an autoregressive manner with causal masking, taking 21 clinical variables, 17 multi-binary treatment actions, and a time-to-next-visit feature as inputs. It predicts the patient's future state across both continuous biomarkers (e.g., tau, amyloid-$\beta$, and brain volume) and cognitive assessments (e.g., ADNI-Mem and ADNI-EF).

\subsubsection{Training Data}

\paragraph{Raw ADNI Dataset}
The data used in the preparation of this article were obtained from the Alzheimer’s Disease Neuroimaging Initiative database (adni.loni.usc.edu). The ADNI was launched in 2003 as a public-private partnership, led by the Principal Investigator Michael W. Weiner, MD. The primary goal of the ADNI was to test whether serial magnetic resonance imaging (MRI), positron emission tomography (PET), other biological markers, and clinical and neuropsychological assessments can be combined to measure the progression of mild cognitive impairment (MCI) and early Alzheimer’s disease.

\paragraph{Longitudinal Data Processing}
Raw longitudinal data from the ADNI dataset were transformed into a structured format that was optimized for autoregressive visit-to-visit state forecasting. Medication records were consolidated into therapeutic drug classes (e.g., AD treatments, statins, and antihypertensives), with activity windows determined by the start and end dates relative to each visit. The partitioning for drug-class mapping can be found in the \nameref{appendix:drug_class_mapping} section of Appendix A. 

Patient age was dynamically recalculated from baseline to ensure temporal consistency, and participants with fewer than three visits were excluded to provide a sufficient trajectory length for modeling. The full outline of the demographic information for the ADNI patient population used to train the CAST model is presented in Table \ref{tab:dataset_summary_full}.

\paragraph{Data Imputation}
Missing data across continuous biomarkers (tau, amyloid-$\beta$, neuroimaging volumes, ADNI-Mem, ADNI-EF, and age) were imputed using an iterative ExtraTrees-based method \cite{scikit-learn}, which was chosen for its ability to capture non-linear relationships without assuming parametric distributions. The imputer and z-score scaler were fitted only to the training set and reused unchanged for validation and testing to avoid data leakage. Categorical variables were one-hot encoded using k–1 dummy coding to prevent multicollinearity.

\paragraph{Resulting Data}
After preprocessing, two datasets were generated: one for the CAST model and the other for the behavior cloning model. The CAST dataset was structured such that each sample represented a patient state–action pair, with the prediction target as the subsequent visit state, which supported the autoregressive loss calculation. A behavior cloning clinician dataset was generated to capture real-world treatment patterns by mapping patient states to prescribed medication regimens, enabling a comparison between the learned agent policies and a learned clinician policy from the dataset. Details on feature and action variables can be found in the Appendix under \nameref{appendix:features_actions} and \nameref{appendix:action_features} respectively.

\paragraph{Dataset Split}
To ensure independence across individuals, the data were split at the subject level into training (70\%), validation (15\%), and test (15\%) sets and were kept the same for the clinician and CAST model datasets.

\begin{table}[t]
\centering
\scriptsize
\caption{Demographic and Trajectory Summary}
\label{tab:dataset_summary_full}
\begin{tabular}{@{}lc@{}}
\toprule
\textbf{Characteristic} & \textbf{Full Dataset} \\
\midrule
\multicolumn{2}{@{}l}{\textit{Overview}} \\
\hspace{1em} Visits & 12{,}984 \\
\hspace{1em} Unique subjects & 1{,}905 \\

\multicolumn{2}{@{}l}{\textit{Demographics}} \\
\hspace{1em} Age (mean $\pm$ SD) & 76.2 $\pm$ 7.4 \\
\hspace{1em} Range (yrs) & 55.0--103.1 \\
\hspace{1em} Sex (Male / Female) & 54.8\% / 45.2\% \\
\hspace{1em} Race (White) & 93.0\% \\
\hspace{1em} Race (Black / Asian / Other) & 3.9\% / 1.7\% / 1.4\% \\

\multicolumn{2}{@{}l}{\textit{Cognitive Scores}} \\
\hspace{1em} ADNI MEM & 0.38 $\pm$ 1.16 \\
\hspace{1em} ADNI EF2 & 0.14 $\pm$ 1.02 \\

\multicolumn{2}{@{}l}{\textit{Trajectory Lengths}} \\
\hspace{1em} Mean visits per subject & 6.82 $\pm$ 3.72 \\
\hspace{1em} Median (IQR) & 6 (4--9) \\
\hspace{1em} Range (min--max) & 3--22 \\
\hspace{1em} 95th percentile & 15 \\

\multicolumn{2}{@{}l}{\textit{Temporal Variable}} \\
\hspace{1em} Next visit interval (months) & 8.6 $\pm$ 6.9 \\

\multicolumn{2}{@{}l}{\textit{Visit Disease State}} \\
\hspace{1em} Healthy / Impaired (MCI/AD) & 71.4\% / 28.6\% \\

\bottomrule
\end{tabular}

\vspace{0.5em}
\footnotesize{Patient demographic information for the ADNI data patients used in the creation of the forecasting model. Impaired visits in this table is measured as ADNI Mem score under -0.1 \cite{Crane_Carle_Gibbons_Insel_Mackin_Gross_Jones_Mukherjee_Curtis_Harvey_et_al._2012}. Distributions across splits were comparable.}
\end{table}

\subsubsection{Training Protocol}
The model was trained to forecast longitudinal patient trajectories using teacher forcing and the AdamW optimizer with an L2 weight decay (0.01). The learning rates were managed using a ReduceLROnPlateau scheduler \cite{paszke2019pytorch}, and regularization was enforced using a dropout rate of 0.3 in the transformer layer.

The composite loss function combined three terms: the mean squared error for continuous variables, binary cross-entropy for categorical outputs, and an auxiliary load-balancing loss (scaled by 0.005) to prevent expert collapse. 

To ensure a robust evaluation, all experiments utilized pre-split training, validation, and test sets with subject identifiers excluded to prevent data from being leaked.

\subsubsection{Forecasting Model Validation}

We validated the CAST model using held-out data to assess its ability to predict the trajectories of unseen patients. Because most ADNI visits occurred at varying time intervals, the CAST model was trained and evaluated based on actual gaps between visits. Validation relied on two complementary tests: the Maximum Mean Discrepancy (MMD) and a Mantel test. The MMD test, which measures the kernel-based distributional similarity between the forecasted and ground truth trajectories, was computed using an RBF kernel (1{,}000 permutations) on z-scored transition vectors. The Mantel test complements the MMD by evaluating whether the pairwise similarity structure between time points (i.e., temporal adjacency in feature space) is preserved between the predicted and ground truth trajectories \cite{Panda2019hyppoAM, Mantel_1967}. The group-level analysis for the Mantel test was split into two tests: a Fisher-Z one-sided t-test on the Mantel r values, and a permutation-based group test. The permutation test used $5{,}000$ permutations, where only the time dimension of the forecasted series was permuted.

This suite of tests was selected because it provides a comprehensive characterization of the dynamic model performance and how it compares with real-world patients.

\paragraph{Evaluation Method} The model was initialized with each test set subject’s first observed state and
generated subsequent transitions using only past predictions and ground truth actions in an autoregressive manner.
This maximized the opportunity for model errors to compound and drift away from the ground truth, thereby faithfully
evaluating the ability of the forecasting model to reconstruct the high-fidelity trajectories.

\paragraph{Results} The trajectory reconstruction in the test set did not show statistically significant differences
in either step-wise dynamics or long-range trends relative to the ground truth trajectories (at $\alpha=0.05$). The short-range MMD, which measures the distribution of one-step visit-level transitions, failed to reject the difference between the forecasted and ground truth trajectories (MMD $= 4.65 \times 10^{-3}$, $p = 0.3916$). Similarly, the long-range MMD, which measures the overall drift from the first to the last visit, failed to reject the difference between the start and end state distributions (MMD $= 3.79 \times 10^{-3}$, $p = 1.00$). At the feature level, no individual feature showed a statistically significant deviation between the two distributions. The Fisher-Z one-sided t-test on Mantel r values showed strong preservation of interfeature dependency between the forecasted and ground truth trajectories in the test set ($r = 0.7104$, $CI= 0.7472 - 0.8102$, $p=1.99\times10^{-4}$, n = 145). The permutation group test (5,000 within-subject timepoint shuffles) showed that the observed mean Mantel-like similarity significantly exceeded the shuffled null hypothesis.

These results demonstrate that the forecasting model captures physiologically valid transition dynamics, generalizes unobserved patient states, and provides a stable foundation for reinforcement learning experiments. The convergence of the MMD and Mantel results indicates a strong distributional and relational similarity between the forecasting model trajectories and ground truth, even in the presence of minor differences in data.

\subsection{ALPACA Environment}
The ALPACA environment wraps the CAST model as its core state transition function.

\subsubsection{State Initialization and Population} To ensure patient privacy no initial patient visit data are used to seed the initial visit within the ALPACA environment. Instead, we trained a Gaussian Mixture Model (GMM) to learn the multivariate distribution of the start states for the ADNI patient cohort. At the beginning of each episode, the environment samples an initial synthetic patient state from the GMM distribution to initialize a virtual patient. To determine the appropriate number of mixture components, we trained the GMMs across a range of cluster counts (1–100) and automatically selected the configuration with the lowest Bayesian Information Criterion (BIC).

To better capture population heterogeneity and allow for more granular cohort selection within the environment, we fitted three separate GMMs using the Scikit-Learn package \cite{scikit-learn}: one using all participants in the ADNI training set, one restricted to cognitively healthy participants, and one restricted to cognitively impaired participants. Consistent with prior work, impairment was defined as having an ADNI-Mem score below –0.1 \cite{Crane_Carle_Gibbons_Insel_Mackin_Gross_Jones_Mukherjee_Curtis_Harvey_et_al._2012}. This approach enabled realistic multivariate sampling of synthetic patient states while maintaining the privacy of individual ADNI participants.

\subsubsection{Action Space and Constraints}
At each time step, the agent selects a subset of multibinary actions that represent the combinations of medication classes. To maintain medical validity, the environment enforces two constraints:
\begin{enumerate}
\item If the agent selects ``No Medication,'' it must not select any other treatments.
\item At least one valid action must be taken.
\end{enumerate}
Violating these constraints results in early termination and a $-10$ reward, encouraging the agents to make explicit and interpretable decisions.

\subsubsection{Transition Dynamics and Validity}
At each step, ALPACA advances the environment by aggregating a patient’s prior visits and autoregressively forecasting the next visit state. During inference, we fixed the transition interval to six months, rather than using irregular visit spacing as in training, to simplify the downstream policy evaluation. After prediction, the states are inverse-transformed using the original feature scaler and validated against the training distribution. Each feature must lie within three standard deviations of the training mean before the state is returned to the agent. If any feature violates this bound, the episode is terminated, and a neutral reward of zero is assigned. This prevents policies from being implicitly penalized or rewarded for CAST model failures and reduces the risk of agents exploiting simulation artifacts to obtain artificially high rewards.

\subsubsection{Reward Function Formulation} The reward signal is centered on ADNI-Mem, a composite cognitive score derived from several clinically evaluated memory tests \cite{Crane_Carle_Gibbons_Insel_Mackin_Gross_Jones_Mukherjee_Curtis_Harvey_et_al._2012}. This metric was chosen for its strong correlation with neuropsychiatric symptom severity, capturing fundamental patient states without being explicitly included in the reward function \cite{De_Vito_Kunicki_Joyce_Huey_Jones_2025}. 

The reward at each step was computed as the change in the ADNI-Mem between consecutive states, weighted by the standard error of the difference (SE). Given the normalized nature of the ADNI-Mem, we set the standard deviation to one. Although a precise test-retest reliability coefficient for the ADNI-Mem composite score has not been reported, a value of $\geq 0.91$ is a standard and conservative expectation for a high-quality neuropsychological composite score used in clinical research. This aligns with the psychometric standards required for measures used as outcome variables in clinical trials. Therefore, a test-retest reliability coefficient of 0.91 was selected as a conservative, evidence-based estimate for calculating the Standard Error of the Difference. 

\noindent The reward $r_{t+1}$ is calculated as
\begin{equation}
\label{eq:reward}
r_{t+1} \;=\; \mathrm{clip}\!\left( 10 \cdot \frac{M_{t+1} - M_t}{M_{\mathrm{diff}}},\ -10,\ 10 \right)
\end{equation}

where:
\begin{itemize}
  \item $M_t$ is the ADNI-Mem score at timestep $t$,
  \item $M_{t+1}$ is the ADNI-Mem value predicted for timestep $t+1$,
  \item $M_{\mathrm{diff}} = \sqrt{2(1 - r_{xx})}\cdot \sigma$ is the standard error of difference,  
        with $\sigma = 1$ for z-scaled ADNI-Mem and $r_{xx}$ the test--retest reliability coefficient,
  \item $\mathrm{clip}(x,\ a,\ b)$ bounds the value of $x$ within the range $[a,\ b]$.
\end{itemize}

\subsubsection{Episode Semantics} Each episode spans up to 22 time steps, representing 11 years of disease progression at six-month intervals. Patient age was incremented accordingly, and each treatment index corresponded to a specific therapeutic class (e.g., statins, antihypertensive agents, and metformin). These design choices ensure that ALPACA produces physiologically valid trajectories and enables an interpretable analysis of the learned policies.

\section{Benchmark Policy Training}

In this section, we propose four RL-based treatment benchmark policies (Proximal Policy Optimization (PPO), Advantage Actor--Critic (A2C), Soft Actor--Critic (SAC), and Branching Dueling Q-Network (BDQ)). We then compared the learned policies with a behavior-cloned clinician policy, optimal treatment heuristic policy, and no medication policy. 

\subsection{Policies}

\subsubsection{RL-based Treatment Policies}
To evaluate ALPACA across diverse reinforcement learning paradigms, we benchmarked four agents: PPO, A2C, SAC, and BDQ. We used Stable Baselines3 \cite{raffin2021stable} for PPO, A2C, and SAC and adapted the official BDQ implementation \cite{Tavakoli2017ActionBA}. Together, these methods span on-policy, off-policy, and value-based learning, providing a broad assessment of agent behavior in medical simulation settings. All models were trained for 500{,}000 time steps across four parallel environments, with input normalization applied to standardize the heterogeneous clinical feature scales.

PPO and A2C used standard two-layer multilayer perceptron (MLP) policies with 256 hidden units per layer. Because the SAC model is formulated for continuous action spaces, we implemented a wrapper that maps discrete medication actions to a continuous box space via dynamic thresholding. This preserves the action interpretability while enabling stable optimization. We additionally stabilized SAC using delayed learning starts and gradient clipping. For BDQ, we used a shared trunk with two 256-unit layers, followed by separate 128-unit branches for each action dimension and a 128-unit value head.

Our analysis focused on patient trajectories corresponding to MCI and AD. Although ALPACA also supports cognitively unimpaired profiles, we prioritized symptomatic patients to better align the evaluation with the clinical objective of treatment optimization.

\begin{figure*}[t]
    \centering
    \includegraphics[width=0.98\linewidth]{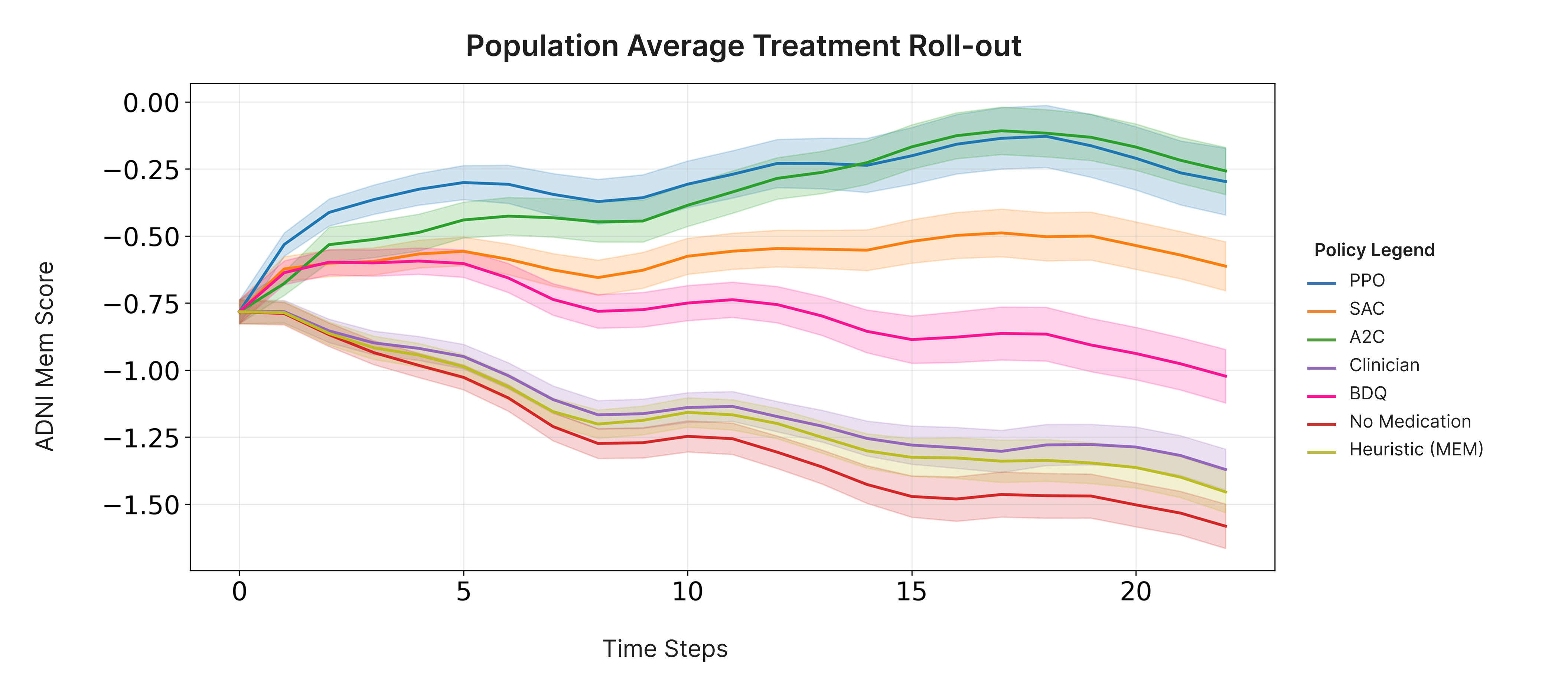}
    \caption{Policy performance over 1000 simulated rollouts (mean and 95\% CI across patients).}

    \label{fig:population_treatment_average}
\end{figure*}

\subsubsection{Clinician Policy}
To provide a real-world baseline for comparison with reinforcement learning agents, we trained a clinician policy Bayesian feedforward neural network (BFNN) using a behavioral cloning approach. This model approximates clinician decision-making by mapping patient states directly to treatment vectors observed in the ADNI dataset. The same preprocessing pipeline and subject splits were used as ALPACA for comparability, with the key distinction that the clinician model was trained to predict multi-binary treatment assignments at each visit, rather than forecasting future states. The input features included 12 continuous and nine categorical patient characteristics, and the output included 17 binary medication indicators that represented different therapeutic classes. During training, we applied a class-balanced binary cross-entropy (BCE) loss with per-action positive weights $(\text{neg} / \text{pos})^{0.55}$ to mitigate class imbalance, while preventing overly aggressive reweighting of rare medication actions. The Bayesian nature of the clinician model architecture was chosen because of its relative simplicity and ability to quantify predictive uncertainty in the evaluation metrics. Because this model serves as a baseline, incorporating uncertainty estimates provides additional context for the traditional performance metrics.

The behavior cloning performance was evaluated using a held-out test set. To assess robustness across multiple draws of patient states, we used Monte Carlo sampling and quantified predictive fidelity using exact-match accuracy and Hamming loss, as well as macro- and micro-averaged F1 scores to account for class imbalance. We also assessed probabilistic calibration using the Brier score and calibration error metrics, including the adaptive calibration error (ACE) and expected calibration error (ECE). All metrics were implemented with open-source Scikit-Learn utilities \cite{scikit-learn}, except ACE and ECE, which followed the implementation described in \cite{Kueppers_2020_CVPR_Workshops}

\subsubsection{Clinician Heuristics Policy}
In addition to the behavior-cloned clinician policy, we implemented a heuristic clinician baseline intended to approximate consensus guideline-based treatment decisions for Alzheimer's disease. This policy follows a simple rule. At each step, it selects the ``No Medication" action unless the patient’s ADNI-Mem score falls below $-0.1$, in which case it selects ``AD Treatment". We chose the $-0.1$ threshold based on the criterion described by Crane et al. for delineating MCI and AD onset using the ADNI-Mem composite score \cite{Crane_Carle_Gibbons_Insel_Mackin_Gross_Jones_Mukherjee_Curtis_Harvey_et_al._2012}. Clinical guidelines recommend initiating AD-specific therapies at the onset of MCI or mild AD \cite{Grossberg2019PresentAA}. Therefore, this heuristic maps our discrete action space onto an idealized treatment strategy grounded in established clinical reasoning while reducing variability and noise relative to behavior-cloned policies trained on observational clinician actions.

\begin{table*}[t]
\centering
\caption{Performance of reinforcement learning agents, no-treatment and behavior cloned clinician baselines evaluated over 1000 simulated patient rollouts. Reported values are means of reward values calculated across rollouts.}
\label{tab:rl_performance}
\begin{tabular}{lccc}
\toprule
\textbf{Policy} & \textbf{Cumulative Reward} & \textbf{Per-step Reward} & \textbf{Final ADNI-Mem} \\
\midrule
No Medication (baseline) & $-17.81$ & $-0.95$ & $-1.53$ \\
Clinician & $-13.72$ & $-0.73$ & $-1.36$ \\
Heuristic & $-15.10$ & $-0.79$ & $-1.42$ \\
A2C & $-0.66$ & $-0.06$ & $-0.59$ \\
PPO & $\mathbf{3.38}$ & $\mathbf{0.27}$ & $\mathbf{-0.46}$ \\
SAC & $0.18$ & $0.01$ & $-0.76$ \\
BDQ & $-7.25$ & $-0.47$ & $-0.98$ \\
\bottomrule
\end{tabular}
\end{table*} 

\subsection{Results}

\subsubsection{Behavior Cloning}
We evaluated the clinician policy on the held-out test set using 50 Monte Carlo simulations. The model achieved an exact match across all 17 medications in $19.9\%$ of cases, but individual medication predictions matched clinician choices in $91\%$ of cases (Hamming loss $=0.090$). This indicates that while the exact treatment combinations were difficult to reproduce, common prescription patterns were captured reliably.

Class imbalance in the dataset posed a significant challenge. The macro F1 score was low ($0.104$), reflecting the under-prediction of rare medications, whereas the micro F1 score was higher ($0.487$), indicating reasonable accuracy at the individual medication level despite skewed action frequencies.

In addition to the point prediction performance, the Bayesian modeling approach enables the assessment of the calibration and uncertainty. The model produced well-calibrated probabilities (Brier $=0.085$, ECE $=0.107$, and ACE $=0.144$) and a strong predictive log-likelihood ($-0.3$). Variance decomposition showed that nearly all uncertainty was aleatoric ($0.133$) rather than epistemic ($0.003$), reflecting the variability among patients rather than model limitations.

Overall, the clinician policy effectively reproduced the dominant prescribing strategies observed in the ADNI while remaining conservative in the use of rare medications. Although it does not perfectly match all clinical actions, its fidelity and calibration make it a robust baseline for comparison with reinforcement learning agents.

\subsubsection{Policy Evaluation Method}
We evaluated all policies on identical cohorts by generating 1000 initial patient states from ALPACA’s start-state model and rolling out each policy from the same starting point. All policies used a normalizing wrapper in StableBaselines3 to normalize the incoming state values. We used the default environment settings and a fixed seed (42) for reproducibility.

\subsubsection{Policy Evaluation Results}
Across seeds, the RL policies significantly outperformed both the no-treatment baseline and the behavior-cloned clinician policy (all $p<0.001$), and the clinician policy outperformed the no-treatment policy ($p<0.001$). All statistical significance was assessed using a non-parametric Wilcoxon within-seed paired test. Taken together, these results support the clinical plausibility of the environment. PPO achieved the strongest average performance (Table~\ref{tab:rl_performance}, Figure~\ref{fig:population_treatment_average}). Although trained policies generally outperform clinician baselines at the group level (Figure~\ref{fig:population_treatment_average}), the outcomes exhibit substantial heterogeneity. In particular, some impaired patients show no improvement relative to clinician policies (Appendix~\ref{appendix:individual_treatment_simulation}).

\begin{figure*}[t]
    \centering
    \includegraphics[width=0.98\textwidth]{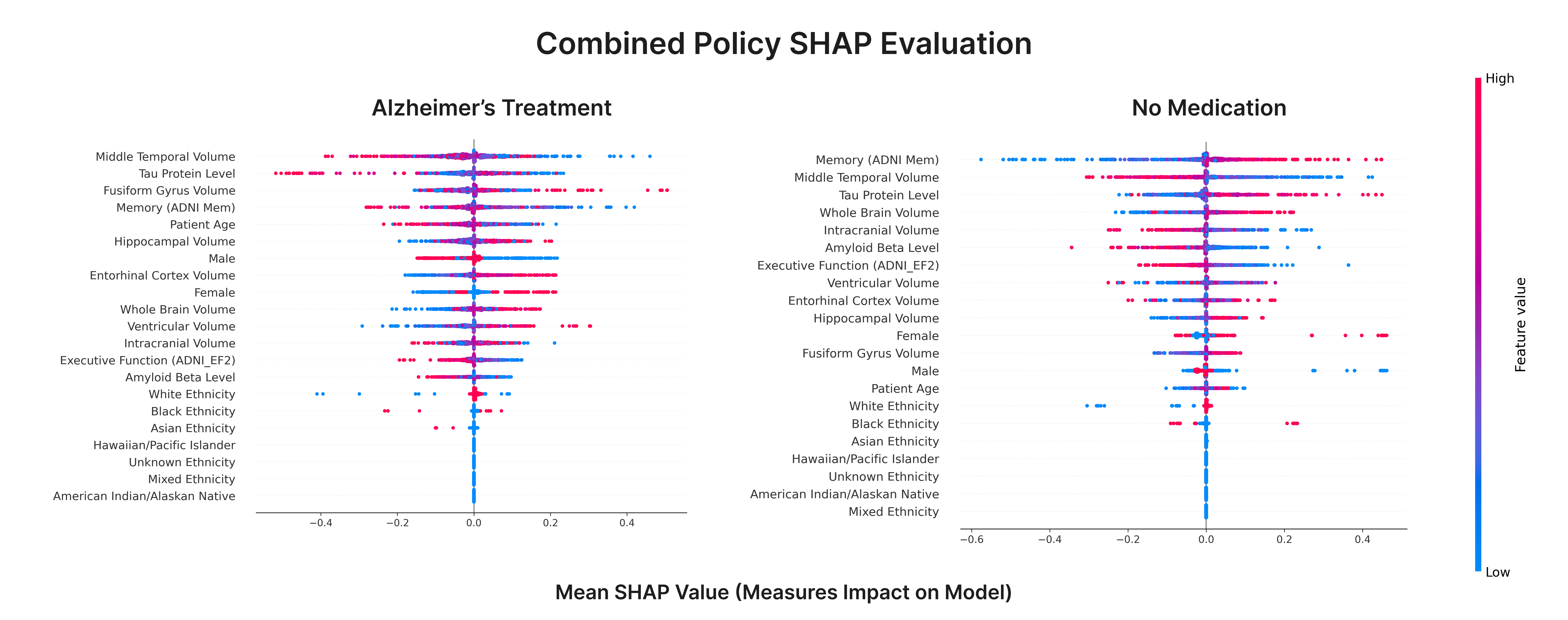}
    \caption{
        Aggregated SHAP values from all models (PPO, SAC, A2C, and BDQ) for the No Medication and AD Treatment actions. The red values indicate higher feature values, and higher SHAP scores indicate a greater propensity to act.
    }
    \label{fig:combined_shap}
\end{figure*}

\subsubsection{SHAP Policy Evaluation}
\label{sec:results}
To evaluate whether the learned policies relied on clinically meaningful signals rather than environment-specific artifacts, we applied SHAP analysis to the policies trained in the ALPACA environment \cite{NIPS2017_7062}. The SHAP values quantify how individual patient features shift the probability of selecting an action. We focus on the \textit{No Medication} and \textit{AD Treatment} actions because they offer the least obscured insights into a policy’s interpretation of disease severity and treatment needs. The results reported here used the aggregated SHAP values across all four trained policies (Figure \ref{fig:combined_shap}).

\paragraph{No Medication}
The attributions for the No Medication action were consistent with clinical expectations. The probability of withholding treatment increased with higher memory scores and relatively preserved whole-brain volume. A higher hippocampal volume similarly increased the likelihood of taking no medical action, consistent with a lower disease burden \cite{Mortimer1997BrainRA, Stern2019BrainRC}. In contrast, elevated amyloid-$\beta$ reduced the likelihood of selecting No Medication. Although the effect magnitudes varied by architecture, the aggregated SHAP patterns converged on these relationships, suggesting that the policies learned a reasonable criterion for when treatment was unnecessary.

\paragraph{AD Treatment}
The attributions for AD Treatment action were also biologically plausible. Higher cognitive performance (memory and executive function) reduced the likelihood of treatment, mirroring the No Medication findings and reflecting a reduced clinical need. In contrast, increased ventricular volume, a structural marker associated with cortical atrophy, increased the probability of selecting AD Treatment. Sex also contributed modestly, with female patients showing a higher tendency toward treatment selection, consistent with the higher prevalence and burden of AD reported in women than in men \cite{Li_Feng_Sun_Hou_Han_Liu_2022}.

\paragraph{Tau and the treatment window}
Higher tau decreases attribution toward ``AD Treatment" and increases attribution toward ``No Medication" (Figure~\ref{fig:combined_shap}). Although this pattern may appear counterintuitive, elevated tau levels are typically associated with later-stage disease and reduced treatment responsiveness. The resulting action attributions suggest that the policy has learned an implicit treatment window, favoring intervention when pathology is more likely to remain modifiable (lower tau) and down-weighting treatment when tau is high and progression is less amenable to treatment. This is consistent with evidence that individuals with high tau levels derive reduced benefits from continued treatment relative to those with lower tau levels \cite{Sims_Zimmer_Evans_Lu_Ardayfio_Sparks_Wessels_Shcherbinin_Wang_Monkul_Nery_et_al._2023}.

Overall, the aggregated SHAP analysis suggests that the learned policies consistently rely on neurobiologically relevant features across architectures, reflecting established markers of Alzheimer’s disease severity and progression, even after only $500{,}000$ training steps. Together, these patterns support the conclusion that ALPACA provides clinically meaningful learning signals, rather than opportunities for reward hacking.

\section{Discussion} 
\label{sec:discussion}

We demonstrated that ALPACA provides a stable foundation for reinforcement learning-based treatment optimization in Alzheimer’s disease. The CAST dynamics model generated realistic medication-conditioned trajectories, supported by MMD and Mantel test analyses, and the aggregate SHAP results suggested that the learned policies were driven by clinically coherent signals rather than reward-hacking artifacts. Together, these findings position ALPACA as a scalable in silico testbed for studying personalized treatment strategies in controlled settings.

The clinical literature suggests that medication effects on disease progression are complex and may vary across individuals, yet these relationships can be obscured when trials estimate average efficacy across heterogeneous populations. We hypothesize that inconsistent findings in repurposing studies may partially reflect unobserved subtype-specific effects that reduce statistical power under conventional trial designs. ALPACA helps address this gap by enabling a systematic in silico exploration of medication regimens across large cohorts of synthetic patients. Although such simulations cannot establish causality, they facilitate controlled hypothesis generation and mechanistic interrogation across diverse disease profiles, helping to prioritize and refine questions for downstream empirical evaluation.

Furthermore, ALPACA offers a valuable testbed for advancing explainable reinforcement learning (XRL) in healthcare. Existing methods, such as SHAP and LIME, often struggle with higher-order dependencies involving polypharmacy and longitudinal dynamics. ALPACA enables the development of medically grounded interpretability techniques that better reflect real-world clinical decision-making. Coupling this environment with expressive XRL methodologies could aid in clarifying targeted hypotheses and narrowing research questions for future experimental validation.

Finally, ALPACA serves as a hypothesis-generating tool for clinical trial planning. By grouping medications according to their mechanism of action, researchers can probe therapeutic behaviors across phenotypes to inform patient stratification and identify potential interference before trial initiation. Intended to complement rather than replace traditional designs, ALPACA provides a reproducible and low-risk setting to enhance the efficiency, precision, and interpretability of neurodegenerative disease research.

\section{Conclusion}
\label{sec:conclusion}
We introduce ALPACA, an open-source, Gym-compatible reinforcement learning environment for Alzheimer’s disease treatment research, driven by a medication-conditioned forecasting model trained on longitudinal trajectories from ADNI. We show that policies trained within ALPACA yield improved memory-related outcomes relative to clinician-derived baselines while remaining consistent with biologically and clinically plausible treatment patterns. We positioned ALPACA as an initial step toward clinically grounded AD simulation environments. The current limitations include the use of discrete medication actions, coarse temporal resolution of states and actions, and limited granularity of patient representations. These constraints motivate future extensions, including continuous action spaces and richer multimodal state representations.

\acks{
We are very grateful to Yu-Yun Tseng for assistance with figures, and to both Yu-Chee Tseng and Yu-Yun Tseng for their valuable feedback on the manuscript.

Data collection and sharing for this project was funded by the Alzheimer's Disease Neuroimaging Initiative
(ADNI) (National Institutes of Health Grant U01 AG024904) and DOD ADNI (Department of Defense award
number W81XWH-12-2-0012). ADNI is funded by the National Institute on Aging, the National Institute of
Biomedical Imaging and Bioengineering, and through generous contributions from the following: AbbVie,
Alzheimer’s Association, Alzheimer’s Drug Discovery Foundation, Araclon Biotech, and BioClinica Inc.; Biogen;
Bristol-Myers Squibb Company; CereSpir, Inc.; Cogstate; Eisai Inc.; Elan Pharmaceuticals, Inc.; Eli Lilly and
Company; EuroImmun; F. Hoffmann-La Roche Ltd and its affiliated company Genentech, Inc.; Fujirebio; GE
Healthcare; IXICO Ltd.; Janssen Alzheimer Immunotherapy Research \& Development, LLC.; Johnson \&
Johnson Pharmaceutical Research \& Development LLC.; Lumosity; Lundbeck; Merck \& Co., Inc.; Meso
Scale Diagnostics, LLC.; NeuroRx Research; Neurotrack Technologies; Novartis Pharmaceuticals
Corporation; Pfizer Inc.; Piramal Imaging; Servier; Takeda Pharmaceutical Company; and Transition
Therapeutics. The Canadian Institutes of Health Research is providing funds to support ADNI clinical sites
in Canada. Private sector contributions are facilitated by the Foundation for the National Institutes of Health
(www.fnih.org). The grantee organization is the Northern California Institute for Research and Education,
and the study is coordinated by the Alzheimer’s Therapeutic Research Institute at the University of Southern
California. ADNI data are disseminated by the Laboratory for Neuro Imaging at the University of Southern California.
}

\bibliography{chil-sample}

@InProceedings{Kueppers_2020_CVPR_Workshops,
   author = {Küppers, Fabian and Kronenberger, Jan and Shantia, Amirhossein and Haselhoff, Anselm},
   title = {Multivariate Confidence Calibration for Object Detection},
   booktitle = {The IEEE/CVF Conference on Computer Vision and Pattern Recognition (CVPR) Workshops},
   month = {June},
   year = {2020}
}

@article{Bossens2022LowVO,
  title={Low Variance Off-policy Evaluation with State-based Importance Sampling},
  author={David M. Bossens and Philip S. Thomas},
  journal={2024 IEEE Conference on Artificial Intelligence (CAI)},
  year={2022},
  pages={871-883},
  url={https://api.semanticscholar.org/CorpusID:254408540}
}

@article{Uehara2022ARO,
  title={A Review of Off-Policy Evaluation in Reinforcement Learning},
  author={Masatoshi Uehara and Chengchun Shi and Nathan Kallus},
  journal={ArXiv},
  year={2022},
  volume={abs/2212.06355},
  url={https://api.semanticscholar.org/CorpusID:254591267}
}

@article{Huang2024SmartPR,
  title={Smart pain relief: Harnessing conservative Q learning for personalized and dynamic pain management},
  author={Yong Huang and Rui Cao and Thomas Hughes and Amir Rahmani},
  journal={Smart Health},
  year={2024},
  url={https://api.semanticscholar.org/CorpusID:273172715}
}

@article{Li_Feng_Sun_Hou_Han_Liu_2022, title={Global, regional, and national burden of Alzheimer’s disease and other dementias, 1990–2019}, volume={14}, ISSN={1663-4365}, url={https://www.frontiersin.orghttps://www.frontiersin.org/journals/aging-neuroscience/articles/10.3389/fnagi.2022.937486/full}, DOI={10.3389/fnagi.2022.937486}, abstractNote={Background
With the increase in the aging population worldwide, Alzheimer’s disease has become a rapidly increasing public health concern. Monitoring the dementia disease burden will support health development strategies by providing scientific data.

Methods
Based on the data obtained from the 2019 Global Burden of Disease (GBD) database, the numbers and age-standardized rates (ASRs) of incidence, prevalence, death, and disability-adjusted life-years (DALYs) of Alzheimer’s disease and other dementias from 1990 to 2019 were analyzed. Calculated estimated annual percentage changes (EAPCs) and Joinpoint regression analyses were performed to evaluate the trends during this period. We also evaluated the correlations between the epidemiology and the sociodemographic index (SDI), an indicator to evaluate the level of social development in a country or region considering the education rate, economic situation, and total fertility rate.

Results
From 1990 to 2019, the incidence and prevalence of Alzheimer’s disease and other dementias increased by 147.95 and 160.84%, respectively. The ASR of incidence, prevalence, death, and DALYs in both men and women consistently increased over the study period. All the ASRs in women were consistently higher than those in men, but the increases were more pronounced in men. In addition, the ASRs of incidence, prevalence, and DALYs were positively correlated with the SDI. Moreover, the proportion of patients over 70 years old with dementia was also positively correlated with the SDI level. Smoking was a major risk factor for the disease burden of dementia in men, while obesity was the major risk factor for women.

Conclusion
From 1990 to 2019, the Alzheimer’s disease burden increased worldwide. This trend was more serious in high-SDI areas, especially among elderly populations in high-SDI areas, who should receive additional attention. Policy-makers should take steps to reverse this situation. Notably, women were at a higher risk for the disease, but the risk in men showed a faster increase. We should give attention to the aging population, attach importance to interventions targeting dementia risk factors, and formulate action plans to address the increasing incidence of dementia.}, journal={Frontiers in Aging Neuroscience}, publisher={Frontiers}, author={Li, Xue and Feng, Xiaojin and Sun, Xiaodong and Hou, Ningning and Han, Fang and Liu, Yongping}, year={2022}, month=oct, language={English} }

@article{Lim2024OMGRLOfflineMG,
  title={OMG-RL:Offline Model-based Guided Reward Learning for Heparin Treatment},
  author={Yooseok Lim and Sujee Lee},
  journal={ArXiv},
  year={2024},
  volume={abs/2409.13299},
  url={https://api.semanticscholar.org/CorpusID:272770561}
}

@article{Sun2022AdversarialRL,
  title={Adversarial reinforcement learning for dynamic treatment regimes},
  author={Zhaohong Sun and Wei Dong and Haomin Li and Zhengxing Huang},
  journal={Journal of biomedical informatics},
  year={2022},
  pages={
          104244
        },
  url={https://api.semanticscholar.org/CorpusID:253652461}
}

@article{Jayaraman2024APO,
  title={A Primer on Reinforcement Learning in Medicine for Clinicians},
  author={Pushkala Jayaraman and Jacob Desman and Moein Sabounchi and Girish N. Nadkarni and Ankit Sakhuja},
  journal={NPJ Digital Medicine},
  year={2024},
  volume={7},
  url={https://api.semanticscholar.org/CorpusID:274302999}
}

@inproceedings{Fujimoto2018OffPolicyDR,
  title={Off-Policy Deep Reinforcement Learning without Exploration},
  author={Scott Fujimoto and David Meger and Doina Precup},
  booktitle={International Conference on Machine Learning},
  year={2018},
  url={https://api.semanticscholar.org/CorpusID:54457299}
}

@article{Levine2020OfflineRL,
  title={Offline Reinforcement Learning: Tutorial, Review, and Perspectives on Open Problems},
  author={Sergey Levine and Aviral Kumar and G. Tucker and Justin Fu},
  journal={ArXiv},
  year={2020},
  volume={abs/2005.01643},
  url={https://api.semanticscholar.org/CorpusID:218486979}
}

@article{Yu2020MOPOMO,
  title={MOPO: Model-based Offline Policy Optimization},
  author={Tianhe Yu and Garrett Thomas and Lantao Yu and Stefano Ermon and James Y. Zou and Sergey Levine and Chelsea Finn and Tengyu Ma},
  journal={ArXiv},
  year={2020},
  volume={abs/2005.13239},
  url={https://api.semanticscholar.org/CorpusID:218900501}
}

@article{Man2014TheUT,
  title={The UVA/PADOVA Type 1 Diabetes Simulator},
  author={Chiara Dalla Man and Francesco Micheletto and Dayu Lv and Marc D. Breton and Boris P. Kovatchev and Claudio Cobelli},
  journal={Journal of Diabetes Science and Technology},
  year={2014},
  volume={8},
  pages={26 - 34},
  url={https://api.semanticscholar.org/CorpusID:33955269}
}

@inproceedings{Oberst2019CounterfactualOE,
  title={Counterfactual Off-Policy Evaluation with Gumbel-Max Structural Causal Models},
  author={Michael Oberst and David A. Sontag},
  booktitle={International Conference on Machine Learning},
  year={2019},
  url={https://api.semanticscholar.org/CorpusID:155092754}
}

@article{Ahn2011DrugSO,
  title={Drug scheduling of cancer chemotherapy based on natural actor-critic approach},
  author={Inkyung Ahn and Jooyoung Park},
  journal={Bio Systems},
  year={2011},
  volume={106 2-3},
  pages={
          121-9
        },
  url={https://api.semanticscholar.org/CorpusID:20937389}
}

@article{Ghaffari2016AMR,
  title={A mixed radiotherapy and chemotherapy model for treatment of cancer with metastasis},
  author={Ali Ghaffari and B. Bahmaie and Mostafa Nazari},
  journal={Mathematical Methods in the Applied Sciences},
  year={2016},
  volume={39},
  pages={4603 - 4617},
  url={https://api.semanticscholar.org/CorpusID:124440566}
}

@article{Luo2024DTRBenchAI,
  title={DTR-Bench: An in silico Environment and Benchmark Platform for Reinforcement Learning Based Dynamic Treatment Regime},
  author={Zhiyao Luo and Mingcheng Zhu and Fenglin Liu and Jiali Li and Yangchen Pan and Jiandong Zhou and Tingting Zhu},
  journal={ArXiv},
  year={2024},
  volume={abs/2405.18610},
  url={https://api.semanticscholar.org/CorpusID:270094986}
}

@article{Wang2023OptimizedGC,
  title={Optimized glycemic control of type 2 diabetes with reinforcement learning: a proof-of-concept trial},
  author={Guangyu Wang and Xiaohong Liu and Zhen Ying and Guoxing Yang and Zhiwei Chen and Zhiwen Liu and Min Zhang and Hongmei Yan and Yuxing Lu and Yuanxu Gao and Kanmin Xue and Xiaoying Li and Ying Chen},
  journal={Nature Medicine},
  year={2023},
  volume={29},
  pages={2633 - 2642},
  url={https://api.semanticscholar.org/CorpusID:261884154}
}

@article{Sims_Zimmer_Evans_Lu_Ardayfio_Sparks_Wessels_Shcherbinin_Wang_Monkul_Nery_et_al._2023, title={Donanemab in Early Symptomatic Alzheimer Disease: The TRAILBLAZER-ALZ 2 Randomized Clinical Trial}, volume={330}, ISSN={0098-7484}, DOI={10.1001/jama.2023.13239}, abstractNote={There are limited efficacious treatments for Alzheimer disease.To assess efficacy and adverse events of donanemab, an antibody designed to clear brain amyloid plaque.Multicenter (277 medical research centers/hospitals in 8 countries), randomized, double-blind, placebo-controlled, 18-month phase 3 trial that enrolled 1736 participants with early symptomatic Alzheimer disease (mild cognitive impairment/mild dementia) with amyloid and low/medium or high tau pathology based on positron emission tomography imaging from June 2020 to November 2021 (last patient visit for primary outcome in April 2023).Participants were randomized in a 1:1 ratio to receive donanemab (n=860) or placebo (n=876) intravenously every 4 weeks for 72 weeks. Participants in the donanemab group were switched to receive placebo in a blinded manner if dose completion criteria were met.The primary outcome was change in integrated Alzheimer Disease Rating Scale (iADRS) score from baseline to 76 weeks (range, 0-144; lower scores indicate greater impairment). There were 24 gated outcomes (primary, secondary, and exploratory), including the secondary outcome of change in the sum of boxes of the Clinical Dementia Rating Scale (CDR-SB) score (range, 0-18; higher scores indicate greater impairment). Statistical testing allocated α of .04 to testing low/medium tau population outcomes, with the remainder (.01) for combined population outcomes.Among 1736 randomized participants (mean age, 73.0 years; 996 [57.4%] women; 1182 [68.1%] with low/medium tau pathology and 552 [31.8%] with high tau pathology), 1320 (76%) completed the trial. Of the 24 gated outcomes, 23 were statistically significant. The least-squares mean (LSM) change in iADRS score at 76 weeks was −6.02 (95% CI, −7.01 to −5.03) in the donanemab group and −9.27 (95% CI, −10.23 to −8.31) in the placebo group (difference, 3.25 [95% CI, 1.88-4.62]; P &lt; .001) in the low/medium tau population and −10.2 (95% CI, −11.22 to −9.16) with donanemab and −13.1 (95% CI, −14.10 to −12.13) with placebo (difference, 2.92 [95% CI, 1.51-4.33]; P &lt; .001) in the combined population. LSM change in CDR-SB score at 76 weeks was 1.20 (95% CI, 1.00-1.41) with donanemab and 1.88 (95% CI, 1.68-2.08) with placebo (difference, −0.67 [95% CI, −0.95 to −0.40]; P &lt; .001) in the low/medium tau population and 1.72 (95% CI, 1.53-1.91) with donanemab and 2.42 (95% CI, 2.24-2.60) with placebo (difference, −0.7 [95% CI, −0.95 to −0.45]; P &lt; .001) in the combined population. Amyloid-related imaging abnormalities of edema or effusion occurred in 205 participants (24.0%; 52 symptomatic) in the donanemab group and 18 (2.1%; 0 symptomatic during study) in the placebo group and infusion-related reactions occurred in 74 participants (8.7%) with donanemab and 4 (0.5%) with placebo. Three deaths in the donanemab group and 1 in the placebo group were considered treatment related.Among participants with early symptomatic Alzheimer disease and amyloid and tau pathology, donanemab significantly slowed clinical progression at 76 weeks in those with low/medium tau and in the combined low/medium and high tau pathology population.ClinicalTrials.gov Identifier: NCT04437511}, number={6}, journal={JAMA}, author={Sims, John R. and Zimmer, Jennifer A. and Evans, Cynthia D. and Lu, Ming and Ardayfio, Paul and Sparks, JonDavid and Wessels, Alette M. and Shcherbinin, Sergey and Wang, Hong and Monkul Nery, Emel Serap and Collins, Emily C. and Solomon, Paul and Salloway, Stephen and Apostolova, Liana G. and Hansson, Oskar and Ritchie, Craig and Brooks, Dawn A. and Mintun, Mark and Skovronsky, Daniel M. and TRAILBLAZER-ALZ 2 Investigators}, year={2023}, month=aug, pages={512–527} }

@article{Tijms_Vromen_Mjaavatten_Holstege_Reus_van_der_Lee_Wesenhagen_Lorenzini_Vermunt_Venkatraghavan_et_al._2024, title={Cerebrospinal fluid proteomics in patients with Alzheimer’s disease reveals five molecular subtypes with distinct genetic risk profiles}, volume={4}, rights={2024 The Author(s)}, ISSN={2662-8465}, DOI={10.1038/s43587-023-00550-7}, abstractNote={Alzheimer’s disease (AD) is heterogenous at the molecular level. Understanding this heterogeneity is critical for AD drug development. Here we define AD molecular subtypes using mass spectrometry proteomics in cerebrospinal fluid, based on 1,058 proteins, with different levels in individuals with AD (n = 419) compared to controls (n = 187). These AD subtypes had alterations in protein levels that were associated with distinct molecular processes: subtype 1 was characterized by proteins related to neuronal hyperplasticity; subtype 2 by innate immune activation; subtype 3 by RNA dysregulation; subtype 4 by choroid plexus dysfunction; and subtype 5 by blood–brain barrier impairment. Each subtype was related to specific AD genetic risk variants, for example, subtype 1 was enriched with TREM2 R47H. Subtypes also differed in clinical outcomes, survival times and anatomical patterns of brain atrophy. These results indicate molecular heterogeneity in AD and highlight the need for personalized medicine.}, number={1}, journal={Nature Aging}, publisher={Nature Publishing Group}, author={Tijms, Betty M. and Vromen, Ellen M. and Mjaavatten, Olav and Holstege, Henne and Reus, Lianne M. and van der Lee, Sven and Wesenhagen, Kirsten E. J. and Lorenzini, Luigi and Vermunt, Lisa and Venkatraghavan, Vikram and Tesi, Niccoló and Tomassen, Jori and den Braber, Anouk and Goossens, Julie and Vanmechelen, Eugeen and Barkhof, Frederik and Pijnenburg, Yolande A. L. and van der Flier, Wiesje M. and Teunissen, Charlotte E. and Berven, Frode S. and Visser, Pieter Jelle}, year={2024}, month=jan, pages={33–47}, language={en} }

@article{Tavakoli2017ActionBA,
  title={Action Branching Architectures for Deep Reinforcement Learning},
  author={Arash Tavakoli and Fabio Pardo and Petar Kormushev},
  journal={ArXiv},
  year={2017},
  volume={abs/1711.08946},
  url={https://api.semanticscholar.org/CorpusID:962757}
}

@article{Grossberg2019PresentAA,
  title={Present Algorithms and Future Treatments for Alzheimer’s Disease},
  author={George T. Grossberg and Gary Tong and Anna D. Burke and Pierre N. Tariot},
  journal={Journal of Alzheimer's Disease},
  year={2019},
  volume={67},
  pages={1157 - 1171},
  url={https://api.semanticscholar.org/CorpusID:73419065}
}

@article{De_Vito_Kunicki_Joyce_Huey_Jones_2025, title={Parallel changes in cognition, neuropsychiatric symptoms, and amyloid in cognitively unimpaired older adults and those with mild cognitive impairment}, volume={21}, ISSN={1552-5260}, DOI={10.1002/alz.14568}, abstractNote={INTRODUCTION
Alzheimer’s disease (AD) diagnosis centers on cognitive impairment despite other early indicators like neuropsychiatric symptoms (NPSs) and amyloid beta (Aβ) accumulation. This study examined how cognition, NPS, and Aβ changes are interrelated over time in individuals without dementia.

METHODS
Participants were 1247 individuals from the Alzheimer’s Disease Neuroimaging Initiative (ADNI)‐2 and ‐3 cohorts with at least 48 months of follow‐up. Cognitive domains were assessed via ADNI composite measures, NPS via the neuropsychiatric inventory, and Aβ via standardized uptake value ratio (SUVR) composite scores. Co‐occurring changes were evaluated with parallel process models.

RESULTS
NPS was longitudinally associated with performance in each cognitive domain. Negative baseline Aβ‐cognition associations were observed in three cognitive domains. No Aβ‐NPS associations were observed.

DISCUSSION
This study demonstrated strong longitudinal relationships between NPS and cognition in preclinical and prodromal stages of AD. Future studies should incorporate NPS into models of disease trajectories to improve early detection and prediction of disease progression.

Highlights

Co‐occurring changes in Aβ, cognition, and neuropsychiatric symptoms are understudied.We found relationships between neuropsychiatric symptoms and cognition.We found baseline, but not longitudinal, Aβ and cognition associations.Changes in neuropsychiatric symptoms should be included in early detection models of ADRD.}, number={2}, journal={Alzheimer’s \& Dementia}, author={De Vito, Alyssa N. and Kunicki, Zachary J. and Joyce, Hannah E. and Huey, Edward D. and Jones, Richard N.}, year={2025}, month=feb, pages={e14568} }

@article{Bhattarai_Rajaganapathy_Das_Kim_Chen_Alzheimers_Disease_Neuroimaging_Initiative_Australian_Imaging_Biomarkers_and_Lifestyle_Flagship_Study_of_Ageing_Dai_Li_Jiang_et_al._2023, title={Using artificial intelligence to learn optimal regimen plan for Alzheimer’s disease}, volume={30}, ISSN={1527-974X}, DOI={10.1093/jamia/ocad135}, abstractNote={BACKGROUND: Alzheimer’s disease (AD) is a progressive neurological disorder with no specific curative medications. Sophisticated clinical skills are crucial to optimize treatment regimens given the multiple coexisting comorbidities in the patient population.
OBJECTIVE: Here, we propose a study to leverage reinforcement learning (RL) to learn the clinicians’ decisions for AD patients based on the longitude data from electronic health records.
METHODS: In this study, we selected 1736 patients from the Alzheimer’s Disease Neuroimaging Initiative (ADNI) database. We focused on the two most frequent concomitant diseases-depression, and hypertension, thus creating 5 data cohorts (ie, Whole Data, AD, AD-Hypertension, AD-Depression, and AD-Depression-Hypertension). We modeled the treatment learning into an RL problem by defining states, actions, and rewards. We built a regression model and decision tree to generate multiple states, used six combinations of medications (ie, cholinesterase inhibitors, memantine, memantine-cholinesterase inhibitors, hypertension drugs, supplements, or no drugs) as actions, and Mini-Mental State Exam (MMSE) scores as rewards.
RESULTS: Given the proper dataset, the RL model can generate an optimal policy (regimen plan) that outperforms the clinician’s treatment regimen. Optimal policies (ie, policy iteration and Q-learning) had lower rewards than the clinician’s policy (mean -3.03 and -2.93 vs. -2.93, respectively) for smaller datasets but had higher rewards for larger datasets (mean -4.68 and -2.82 vs. -4.57, respectively).
CONCLUSIONS: Our results highlight the potential of using RL to generate the optimal treatment based on the patients’ longitude records. Our work can lead the path towards developing RL-based decision support systems that could help manage AD with comorbidities.}, number={10}, journal={Journal of the American Medical Informatics Association: JAMIA}, author={Bhattarai, Kritib and Rajaganapathy, Sivaraman and Das, Trisha and Kim, Yejin and Chen, Yongbin and Alzheimer’s Disease Neuroimaging Initiative and Australian Imaging Biomarkers and Lifestyle Flagship Study of Ageing and Dai, Qiying and Li, Xiaoyang and Jiang, Xiaoqian and Zong, Nansu}, year={2023}, month=sept, pages={1645–1656}, language={eng} }

@article{DBLP:journals/corr/ShazeerMMDLHD17,
  author       = {Noam Shazeer and
                  Azalia Mirhoseini and
                  Krzysztof Maziarz and
                  Andy Davis and
                  Quoc V. Le and
                  Geoffrey E. Hinton and
                  Jeff Dean},
  title        = {Outrageously Large Neural Networks: The Sparsely-Gated Mixture-of-Experts
                  Layer},
  journal      = {CoRR},
  volume       = {abs/1701.06538},
  year         = {2017},
  url          = {http://arxiv.org/abs/1701.06538},
  eprinttype    = {arXiv},
  eprint       = {1701.06538},
  bibsource    = {dblp computer science bibliography, https://dblp.org}
}

@article{Mantel_1967, title={The Detection of Disease Clustering and a Generalized Regression Approach}, volume={27}, ISSN={0008-5472}, abstractNote={The problem of identifying subtle time-space clustering of disease, as may be occurring in leukemia, is described and reviewed. Published approaches, generally associated with studies of leukemia, not dependent on knowledge of the underlying population for their validity, are directed towards identifying clustering by establishing a relationship between the temporal and the spatial separations for the n(n - 1)/2 possible pairs which can be formed from the n observed cases of disease. Here it is proposed that statistical power can be improved by applying a reciprocal transform to these separations. While a permutational approach can give valid probability levels for any observed association, for reasons of practicability, it is suggested that the observed association be tested relative to its permutational variance. Formulas and computational procedures for doing so are given.While the distance measures between points represent symmetric relationships subject to mathematical and geometric regularities, the variance formula developed is appropriate for arbitrary relationships. Simplified procedures are given for the case of symmetric and skew-symmetric relationships. The general procedure is indicated as being potentially useful in other situations as, for example, the study of interpersonal relationships. Viewing the procedure as a regression approach, the possibility for extending it to nonlinear and multivariate situations is suggested.Other aspects of the problem and of the procedure developed are discussed.Similarly, pure temporal clustering can be identified by a study of incidence rates in periods of widespread epidemics. In point of fact, many epidemics of communicable diseases are somewhat local in nature and so these do actually constitute temporal-spatial clusters. For leukemia and similar diseases in which cases seem to arise substantially at random rather than as clear-cut epidemics, it is necessary to devise sensitive and efficient procedures for detecting any nonrandom component of disease occurrence.Various ingenious procedures which statisticians have developed for the detection of disease clustering are reviewed here. These procedures can be generalized so as to increase their statistical validity and efficiency. The technic to be given below for imparting statistical validity to the procedures already in vogue can be viewed as a generalized form of regression with possible useful application to problems arising in quite different contexts.}, number={2\_Part\_1}, journal={Cancer Research}, author={Mantel, Nathan}, year={1967}, month=feb, pages={209–220} }

@article{paszke2019pytorch,
  title={{PyTorch: An Imperative Style, High-Performance Deep Learning Library}},
  author={Paszke, Adam and Gross, Sam and Chintala, Soumith and Chanan, Gregory and Yang, Edward and DeVito, Zachary and Lin, Zeming and Desmaison, Alban and Antiga, Luca and Lerer, Adam and McDonnell, James and Jia, Zhichao and Zhu, Fan and Liu, Michael and Deng, Xiaowei and Mangalam, Arvind and Singh, Bhargav and Fang, Ye and Lu, Honghao and Sourek, Tero and Kang, Viktor},
  journal={Advances in Neural Information Processing Systems},
  volume={32},
  pages={8024--8035},
  year={2019}
}

@article{Crane_Carle_Gibbons_Insel_Mackin_Gross_Jones_Mukherjee_Curtis_Harvey_et_al._2012, title={Development and assessment of a composite score for memory in the Alzheimer’s Disease Neuroimaging Initiative (ADNI)}, volume={6}, ISSN={1931-7565}, DOI={10.1007/s11682-012-9186-z}, abstractNote={We sought to develop and evaluate a composite memory score from the neuropsychological battery used in the Alzheimer’s Disease (AD) Neuroimaging Initiative (ADNI). We used modern psychometric approaches to analyze longitudinal Rey Auditory Verbal Learning Test (RAVLT, 2 versions), AD Assessment Schedule - Cognition (ADAS-Cog, 3 versions), Mini-Mental State Examination (MMSE), and Logical Memory data to develop ADNI-Mem, a composite memory score. We compared RAVLT and ADAS-Cog versions, and compared ADNI-Mem to RAVLT recall sum scores, four ADAS-Cog-derived scores, the MMSE, and the Clinical Dementia Rating Sum of Boxes. We evaluated rates of decline in normal cognition, mild cognitive impairment (MCI), and AD, ability to predict conversion from MCI to AD, strength of association with selected imaging parameters, and ability to differentiate rates of decline between participants with and without AD cerebrospinal fluid (CSF) signatures. The second version of the RAVLT was harder than the first. The ADAS-Cog versions were of similar difficulty. ADNI-Mem was slightly better at detecting change than total RAVLT recall scores. It was as good as or better than all of the other scores at predicting conversion from MCI to AD. It was associated with all our selected imaging parameters for people with MCI and AD. Participants with MCI with an AD CSF signature had somewhat more rapid decline than did those without. This paper illustrates appropriate methods for addressing the different versions of word lists, and demonstrates the additional power to be gleaned with a psychometrically sound composite memory score.}, number={4}, journal={Brain Imaging and Behavior}, author={Crane, Paul K. and Carle, Adam and Gibbons, Laura E. and Insel, Philip and Mackin, R. Scott and Gross, Alden and Jones, Richard N. and Mukherjee, Shubhabrata and Curtis, S. McKay and Harvey, Danielle and Weiner, Michael and Mungas, Dan and for the Alzheimer’s Disease Neuroimaging Initiative}, year={2012}, month=dec, pages={502–516}, language={en} }

@incollection{NIPS2017_7062,
title = {A Unified Approach to Interpreting Model Predictions},
author = {Lundberg, Scott M and Lee, Su-In},
booktitle = {Advances in Neural Information Processing Systems 30},
editor = {I. Guyon and U. V. Luxburg and S. Bengio and H. Wallach and R. Fergus and S. Vishwanathan and R. Garnett},
pages = {4765--4774},
year = {2017},
publisher = {Curran Associates, Inc.},
url = {http://papers.nips.cc/paper/7062-a-unified-approach-to-interpreting-model-predictions.pdf}
}

@article{Stern2019BrainRC,
  title={Brain reserve, cognitive reserve, compensation, and maintenance: operationalization, validity, and mechanisms of cognitive resilience},
  author={Yaakov Stern and Carol A. Barnes and Cheryl L. Grady and Richard N. Jones and Naftali Raz},
  journal={Neurobiology of Aging},
  year={2019},
  volume={83},
  pages={124-129},
  url={https://api.semanticscholar.org/CorpusID:207973821}
}

@article{Mortimer1997BrainRA,
  title={Brain reserve and the clinical expression of Alzheimer's disease.},
  author={James A. Mortimer},
  journal={Geriatrics},
  year={1997},
  volume={52 Suppl 2},
  pages={
          S50-3
        },
  url={https://api.semanticscholar.org/CorpusID:9810796}
}

@inproceedings{Panda2019hyppoAM,
  title={hyppo: A Multivariate Hypothesis Testing Python Package},
  author={Sambit Panda and Satish Palaniappan and Junhao Xiong and Eric W. Bridgeford and Ronak D. Mehta and Cencheng Shen and Joshua T. Vogelstein},
  year={2019},
  url={https://api.semanticscholar.org/CorpusID:195798646}
}

@article{Sims2023DonanemabIE,
  title={Donanemab in Early Symptomatic Alzheimer Disease: The TRAILBLAZER-ALZ 2 Randomized Clinical Trial.},
  author={John R. Sims and Jennifer A. Zimmer and Cynthia Duggan Evans and Ming-ning Lu and Paul A. Ardayfio and Jondavid Sparks and Alette M. Wessels and Sergey Shcherbinin and Hong Wang and Emel Serap Monkul Nery and Emily C. Collins and Paul R. Solomon and Stephen Salloway and Liana G. Apostolova and Oskar Hansson and Craig W. Ritchie and Dawn A. Brooks and Mark Mintun and Daniel M. Skovronsky},
  journal={JAMA},
  year={2023},
  url={https://api.semanticscholar.org/CorpusID:259946737}
}

@article{Tarawneh2024TheSF,
  title={The search for clarity regarding “clinically meaningful outcomes” in Alzheimer disease clinical trials: CLARITY-AD and Beyond},
  author={Rawan Tarawneh and V. Shane Pankratz},
  journal={Alzheimer's Research \& Therapy},
  year={2024},
  volume={16},
  url={https://api.semanticscholar.org/CorpusID:267701073}
}

@article{Perneczky2025ClinicallyMB,
  title={Clinically meaningful benefit and real‐world evidence in Alzheimer's disease research and care},
  author={Robert Perneczky and Lutz Froelich},
  journal={Alzheimer's \& Dementia : Translational Research \& Clinical Interventions},
  year={2025},
  volume={11},
  url={https://api.semanticscholar.org/CorpusID:278094428}
}

@article{Amirrad_Bousoik_Shamloo_Al-Shiyab_Nguyen_Montazeri_Aliabadi_2017, title={Alzheimer’s Disease: Dawn of a New Era?}, volume={20}, ISSN={1482-1826, 1482-1826}, DOI={10.18433/J3VS8P}, abstractNote={Alzheimer’s disease (AD) is an irreversible neurodegenerative disease characterized by a progressive decline in cognition and memory, leading to significant impairment in daily activities and ultimately death. It is the most common cause of dementia, the prevalence of which increases with age; however, age is not the only predisposing factor. The pathology of this cognitive impairing disease is still not completely understood, which has limited the development of valid therapeutic options. Recent years have witnessed a wide range of novel approaches to combat this disease, so that they greatly increased our understanding of the disease and of the unique drug development issues associated with this disease. In this paper, we provide a brief overview of the history, the clinical presentation and diagnosis, and we undertake a comprehensive review of the various approaches that have been brought to clinical trials in recent years, including immunotherapeutic approaches, tau-targeted strategies, neurotransmitter-based therapies, neurotropic and hematopoietic growth factors, and antioxidant therapies, trying to highlight the lessons learned from these approaches. This article is open to POST-PUBLICATION REVIEW. Registered readers (see “For Readers”) may comment by clicking on ABSTRACT on the issue’s contents page.}, journal={Journal of Pharmacy \& Pharmaceutical Sciences}, author={Amirrad, Farideh and Bousoik, Emira and Shamloo, Kiumars and Al-Shiyab, Hassan and Nguyen, Viet-Huong V. and Montazeri Aliabadi, Hamidreza}, year={2017}, month=jul, pages={184} }

@article{Winblad_Wimo_Engedal_Soininen_Verhey_Waldemar_Wetterholm_Haglund_Zhang_Schindler_2006, title={3-Year Study of Donepezil Therapy in Alzheimer’s Disease: Effects of Early and Continuous Therapy}, volume={21}, rights={https://www.karger.com/Services/SiteLicenses}, ISSN={1420-8008, 1421-9824}, DOI={10.1159/000091790}, abstractNote={Delays in the diagnosis of Alzheimer’s disease, and, therefore, delays in treatment, may have a detrimental effect on a patient’s long-term well-being. This studyassessed the effects of postponing donepezil treatment for 1 year by comparing patients treated continuously for 3 years with those who received placebo for 1 year followed by open-label donepezil for 2 years. Patients (n = 286) with possible or probable Alzheimer’s disease (according to DSM-IV, NINCDS-ADRDA, and Mini-Mental State Examination criteria; see text) were randomized to receive donepezil (5 mg/day for 4 weeks, 10 mg/day thereafter) or placebo (delayed-start group) for 1 year. Of the 192 completers, 157 began a 2-year, open-label phase of donepezil treatment. Outcome measures were the Gottfries-Bråne-Steen scale, the Mini-Mental State Examination, the Global Deterioration Scale, the Progressive Deterioration Scale, the Neuropsychiatric Inventory, and safety (adverse events). Mixed regression analysis was used to compare changes between the groups over 3 years on the efficacy measures. There was a trend for patients receiving continuous therapy to have less global deterioration (Gottfries-Bråne-Steen scale) than those who had delayed treatment (p = 0.056). Small but statistically significant differences between the groups were observed for the secondary measures of cognitive function (Mini-Mental State Examination; p = 0.004) and cognitive and functional abilities (Global Deterioration Scale; p = 0.0231) in favor of continuous donepezil therapy. Over 90% of the patients in both cohorts experienced one treatment-emergent adverse event; most were considered mild or moderate. In conclusion, patients in whom the start of treatment is delayed may demonstrate slightly reduced benefits as compared with those seen in patients starting donepezil therapy early in the course of Alzheimer’s disease. These data support the long-term efficacy and safety of donepezil.}, number={5–6}, journal={Dementia and Geriatric Cognitive Disorders}, author={Winblad, B. and Wimo, A. and Engedal, K. and Soininen, H. and Verhey, F. and Waldemar, G. and Wetterholm, A.-L. and Haglund, A. and Zhang, R. and Schindler, R.}, year={2006}, pages={353–363}, language={en} }

@article{raffin2021stable,
  title={Stable-baselines3: Reliable reinforcement learning implementations},
  author={Raffin, Antonin and Hill, Ashley and Gleave, Adam and Kanervisto, Anssi and Ernestus, Maximilian and Dormann, Noah},
  journal={Journal of machine learning research},
  volume={22},
  number={268},
  pages={1--8},
  year={2021}
}

@article{Saboo_Choudhary_Cao_Worrell_Jones_Iyer_2021, title={Reinforcement Learning based Disease Progression Model for Alzheimer’s Disease}, url={https://www.semanticscholar.org/paper/Reinforcement-Learning-based-Disease-Progression-Saboo-Choudhary/fe2c1b25d4e4bb16612a5381c55f51602943180a}, abstractNote={We model Alzheimer’s disease (AD) progression by combining differential equations (DEs) and reinforcement learning (RL) with domain knowledge. DEs provide relationships between some, but not all, factors relevant to AD. We assume that the missing relationships must satisfy general criteria about the working of the brain, for e.g., maximizing cognition while minimizing the cost of supporting cognition. This allows us to extract the missing relationships by using RL to optimize an objective (reward) function that captures the above criteria. We use our model consisting of DEs (as a simulator) and the trained RL agent to predict individualized 10-year AD progression using baseline (year 0) features on synthetic and real data. The model was comparable or better at predicting 10-year cognition trajectories than state-of-the-art learning-based models. Our interpretable model demonstrated, and provided insights into,"recovery/compensatory"processes that mitigate the effect of AD, even though those processes were not explicitly encoded in the model. Our framework combines DEs with RL for modelling AD progression and has broad applicability for understanding other neurological disorders.}, journal={ArXiv}, author={Saboo, Krishnakant V. and Choudhary, A. and Cao, Yurui and Worrell, G. and Jones, David T. and Iyer, R.}, year={2021}, month=jun }

@article{scikit-learn,
  title={Scikit-learn: Machine Learning in {P}ython},
  author={Pedregosa, F. and Varoquaux, G. and Gramfort, A. and Michel, V.
          and Thirion, B. and Grisel, O. and Blondel, M. and Prettenhofer, P.
          and Weiss, R. and Dubourg, V. and Vanderplas, J. and Passos, A. and
          Cournapeau, D. and Brucher, M. and Perrot, M. and Duchesnay, E.},
  journal={Journal of Machine Learning Research},
  volume={12},
  pages={2825--2830},
  year={2011}
}

\appendix

\clearpage
\section{Patient Features and Actions}
\label{appendix:features_actions}

\subsection{Patient Features}

\begin{compactitem}
    \item \textbf{Cognitive Measures:} \newline ADNI\_MEM, ADNI\_EF2
    \item \textbf{Biomarkers:} \newline TAU\_data, ABETA
    \item \textbf{Demographics:} \newline subject\_age, PTGENDER\_Female, PTGENDER\_Male
    \item \textbf{Race Indicators:} \newline 
    PTRACCAT\_Am Indian/Alaskan, PTRACCAT\_Asian, PTRACCAT\_Black, 
    PTRACCAT\_Hawaiian/Other PI, PTRACCAT\_More than one, 
    PTRACCAT\_Unknown, PTRACCAT\_White
    \item \textbf{Structural MRI Volumes:} \newline 
    Ventricles, Hippocampus, WholeBrain, Entorhinal, 
    Fusiform, MidTemp, ICV
    \item \textbf{Longitudinal Timing:} \newline next\_visit\_months
\end{compactitem}

\subsection{Action Space}
\label{appendix:action_features}

The medication action space consisted of 17 binary indicators denoting whether each medication class was active during a visit.

\begin{compactitem}
    \item AD Treatment\_active
    \item Alpha Blocker\_active
    \item Analgesic\_active
    \item Antidepressant\_active
    \item Antihypertensive\_active
    \item Bone Health\_active
    \item Diabetes Medication\_active
    \item Diuretic\_active
    \item NSAID\_active
    \item No Medication\_active
    \item Other\_active
    \item PPI\_active
    \item SSRI\_active
    \item Statin\_active
    \item Steroid\_active
    \item Supplement\_active
    \item Thyroid Hormone\_active
\end{compactitem}

\clearpage
\section{Drug Class Mapping}
\label{appendix:drug_class_mapping}

The following mapping was used to consolidate individual medications into therapeutic classes for the action space.

\begin{table}[h]
\centering
\caption{Drug–class mapping used in ALPACA.}
\resizebox{\textwidth}{!}{%
\begin{tabular}{ll}
\toprule
\textbf{Drug} & \textbf{Mapped Class} \\
\midrule
Aricept, Donepezil, Namenda, Exelon & AD Treatment \\
Lipitor, Simvastatin, Crestor, Zocor, Atorvastatin & Statin \\
Lisinopril, Atenolol, Amlodipine, Metoprolol, Norvasc, Losartan & Antihypertensive \\
Levothyroxine, Synthroid & Thyroid Hormone \\
Aspirin, Ibuprofen, Aleve, ASA & NSAID \\
Tylenol, Acetaminophen & Analgesic \\
Zoloft, Lexapro, Sertraline, Citalopram, Prozac & SSRI \\
Trazodone & Antidepressant \\
Metformin & Diabetes Medication \\
Vitamin D, Vitamin D3, Vitamin B12, Vitamin C, Vitamin E, Calcium, Multivitamin, Fish Oil & Supplement \\
Omeprazole, Prilosec & PPI \\
Hydrochlorothiazide & Diuretic \\
Fosamax & Bone Health \\
Prednisone, Prednisolone & Steroid \\
Flomax & Alpha Blocker \\
No medication & No Medication \\
\bottomrule
\end{tabular}%
}
\end{table}

\onecolumn
\section{Influence of Patient Feature on Treatment Action}
\label{appendix:alpaca_shap_analysis}

These heatmaps were derived from the SHAP values calculated during inference, as each policy treated patients within the ALPACA environment. The SHAP values measure the relative impact of each patient feature on treatment. A higher score indicates that the feature plays a critical role in the learned policy and utilizes medications in the treatment regimen.

\begin{figure}[h!]
    \centering
    \includegraphics[width=0.98\textwidth]{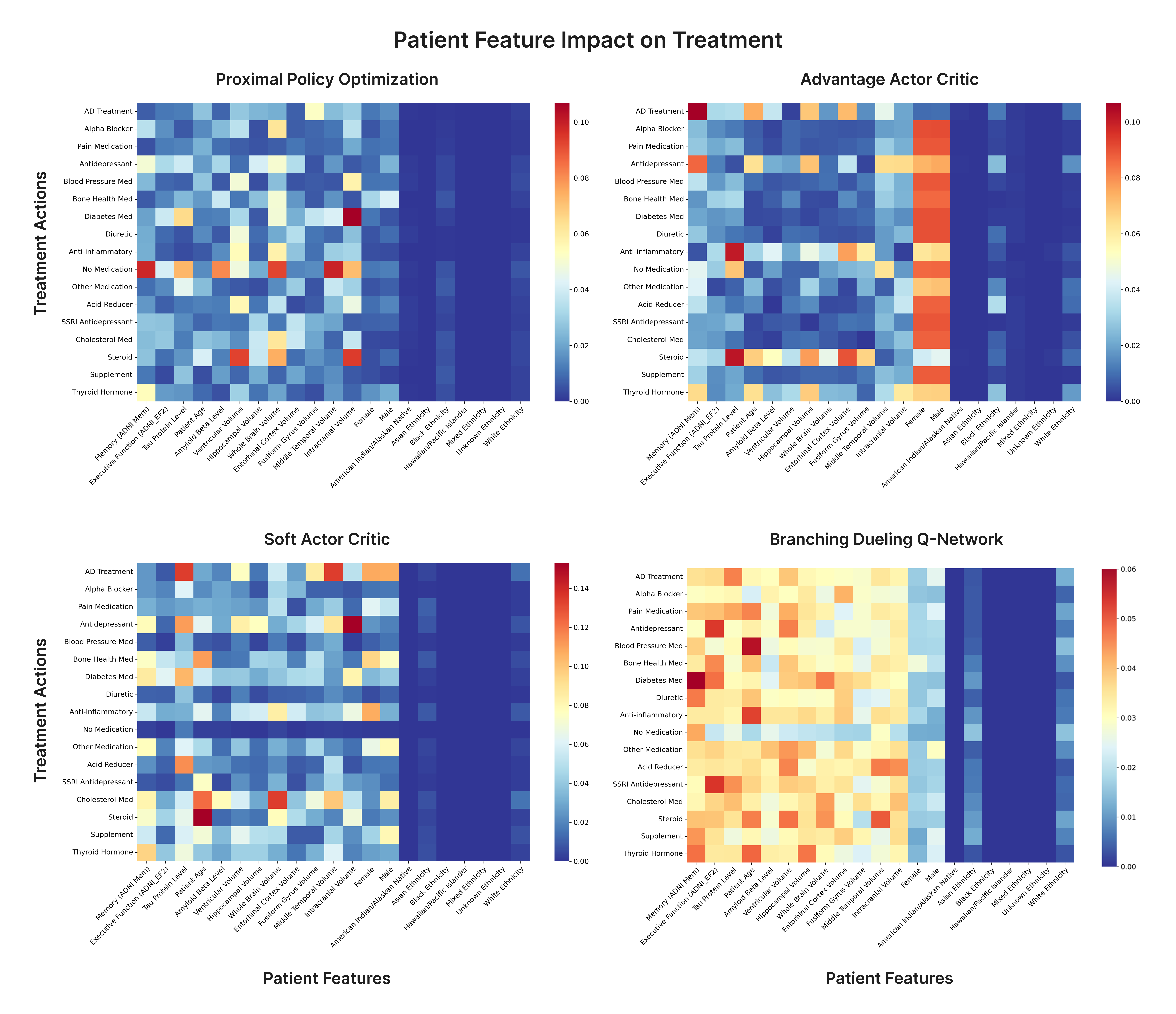}
    \label{fig:feature_action_heatmap}
\end{figure}

\twocolumn

\onecolumn
\section{Individual Treatment Simulation}
\label{appendix:individual_treatment_simulation}

\begin{figure}[H]
    \centering
    \includegraphics[width=0.98\textwidth]{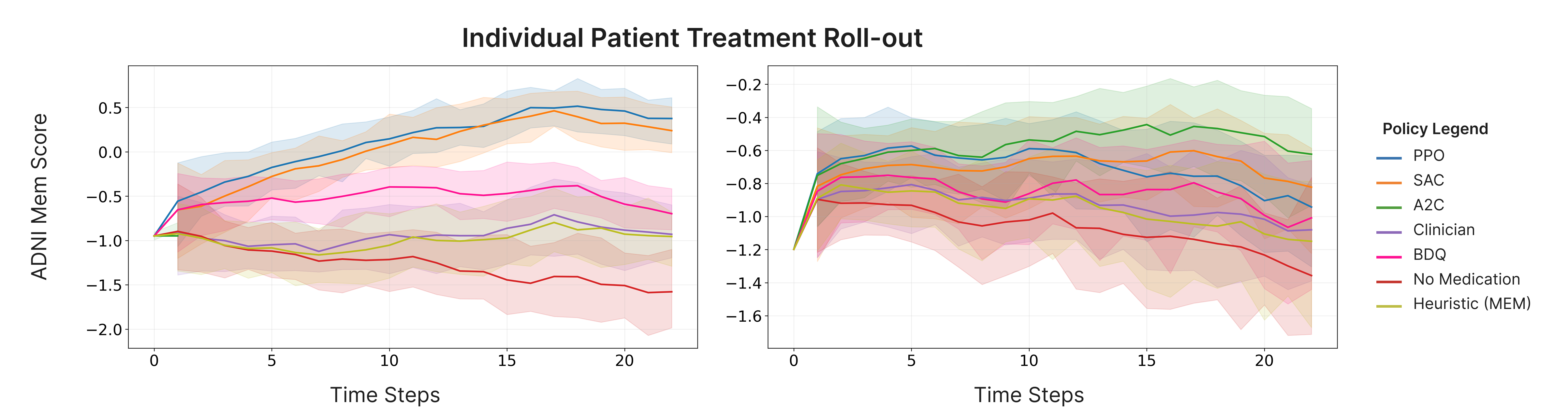}
    \label{fig:patient_rollout}
\end{figure}

This figure shows two synthetic patients generated in the ALPACA environment, illustrating the substantial variability in treatment trajectories among individuals with MCI or AD symptoms. Although policies perform well on average, these examples highlight patient-specific differences that are likely driven by underlying disease heterogeneity. Prior work suggests that Alzheimer's disease comprises multiple biological subtypes, and our Gaussian Mixture Model sampling similarly revealed five to six unsupervised clusters in the ADNI cohort. Because ALPACA initializes patients via weighted subtype sampling, heterogeneity in patient treatment quality may be a function of exposure to patients with various subtypes.

\twocolumn


\end{document}